\documentclass[3p,times,procedia]{elsarticle}
\usepackage{ecrc}
\usepackage{amsmath}
\usepackage[bookmarks=false]{hyperref}
    \hypersetup{colorlinks,
      linkcolor=blue,
      citecolor=blue,
      urlcolor=blue}
\usepackage{amsmath}
\usepackage{tikz}
\usetikzlibrary{arrows,positioning,shadows}
\usepackage{tikz}
\usetikzlibrary{positioning}
\usepackage{enumerate}
\usepackage{amssymb}
\usepackage{pifont}
\newcommand{\cmark}{\ding{51}}%
\newcommand{\xmark}{\ding{55}}%
\usepackage{multirow}
\usepackage{float}
\usepackage{paralist,array}
\usepackage{algorithm}
\usepackage{algorithmic}
\usepackage{graphicx}
\usepackage{lscape}
\usepackage{longtable}
\usepackage{booktabs}
\usepackage{xassoccnt}

\volume{00}

\firstpage{1}

\journalname{Procedia Computer Science}

\runauth{Yury Kaminsky, Irina Deeva}

\jid{procs}

\usepackage{amssymb}

\usepackage[figuresright]{rotating}

\begin{document}
\begin{frontmatter}



\dochead{11th International Young Scientists Conference on Computational Science}

\title{BigBraveBN: algorithm of structural learning for bayesian networks with a large number of nodes}


\author[a]{Yury Kaminsky} 
\author[a]{Irina Deeva\corref{cor1}}

\address[a]{ITMO University, Saint-Petersburg, Russia}

\begin{abstract}
Learning a Bayesian network is an NP-hard problem and with an increase in the number of nodes, classical algorithms for learning the structure of Bayesian networks become inefficient. In recent years, some methods and algorithms for learning Bayesian networks with a high number of nodes (more than 50) were developed. But these solutions have their disadvantages, for instance, they only operate one type of data (discrete or continuous) or their algorithm has been created to meet a specific nature of data (medical, social, etc.). The article presents a $BigBraveBN$ algorithm for learning large Bayesian Networks with a high number of nodes (over 100). The algorithm utilizes the Brave coefficient that measures the mutual occurrence of instances in several groups. To form these groups, we use the method of nearest neighbours based on the Mutual information (MI) measure. In the experimental part of the article, we compare the performance of $BigBraveBN$ to other existing solutions on multiple data sets both discrete and continuous. The experimental part also represents tests on real data. The aforementioned experimental results demonstrate the efficiency of the $BigBraveBN$ algorithm in structure learning of Bayesian Networks.
\end{abstract}

\cortext[cor1]{iriny.deeva@gmail.com}
\end{frontmatter}

\email{iriny.deeva@gmail.com}



\section{Introduction}
\label{intro}
Modelling of real-world objects is often associated with the need to analyze a large number of features, since the more complex the object, the more information about it needs to be taken into account. Thus, when analyzing such complex objects in the real world, a specialist is faced with the need to model multidimensional distributions. Bayesian networks are a convenient tool for modelling multivariate distributions because the sparse structure of the directed graph reduces the number of parameters that need to be estimated. However, if Bayesian networks are used to model real-world objects, it is necessary to solve the problem of learning the Bayesian network structure from data. In this case, the number of possible structures grows superexponentially depending on the number of modelled features (nodes) \cite{chickering1996learning}. In this case, it is necessary to develop learning algorithms for Bayesian networks with a large number of nodes, since there are areas in which the number of simulated features is quite large (more than 100) \cite{liu2016inference}. 

Existing approaches to learning Bayesian networks with a large number of nodes can be divided into two groups - those that limit the search space \cite{scanagatta2015learning} and those that share the task of learning smaller networks with their subsequent connection \cite{liu2016inference}. However, the existing algorithms and methods have several disadvantages. First, they are often designed to work with data of a particular nature, such as medical data or genotype data, where a large number of features is a common situation \cite{michiels2021bayesuites, liu2016inference}. Secondly, the proposed tools for training Bayesian networks with a large number of nodes often work with data of the same type (continuous \cite{aragamlearning} or discrete \cite{suter2021bayesian}). But the objects being modelled are often complex composite probabilistic objects described by features of different types.

Given the above problems, we propose an algorithm that has a high degree of use, that is, it can be applied to data of a different nature, and also effectively works with data of various types (continuous and discrete). The proposed algorithm can be attributed to the family of algorithms that limit the search space, thereby making it possible to train networks with a large number of nodes. The algorithm is based on the Brave coefficient which measures the strength of the connection between features due to the frequencies of their occurrence in different groups. By groups, we mean clusters obtained using the K-nearest neighbours method (KNN) for each feature. Here, we have explored various distance metrics for the KNN algorithm and selected the most efficient ones. Various hyperparameter thresholds were also investigated and conclusions were drawn for the most effective values. The resulting algorithm was compared with the existing analogues of $sparsebn$ and $BiDAG$ and showed the best efficiency in terms of restoring the known structures of Bayesian networks, while its operation time was either less than the time of analogues or comparable.

\begin{nomenclature}
\begin{deflist}[WMO]
\defitem{BN}\defterm{Bayesian network}
\defitem{HC}\defterm{Hill-Climbing}
\defitem{DAG}\defterm{Directed acyclic graph}
\defitem{MI}\defterm{Mutual information}
\defitem{BIC}\defterm{Bayesian information criterion}
\defitem{KNN}\defterm{K-nearest neighbours}
\defitem{SHD} \defterm{Structural Hamming Distance}
\end{deflist}
\end{nomenclature}

\section{Related works}
\label{related}

\subsection{Algorithms for learning Bayesian networks}

A problem of structure learning of large Bayesian Networks is a problem of finding a directed acyclic graph (DAG), that represents relations between features (nodes) as edges. In most real-world situations the problem has a polynomial computational complexity \cite{nagarajan2013bayesian}.

\subsubsection{Search space limiting algorithms}

Search space limiting algorithms utilize different methods to limit the space of possible graphs. Score-based approaches are implemented after limiting solution space in most cases. 

One of the examples of this approach is the MIIGA algorithm. The MIIGA algorithm uses the mutual information (MI) metric \cite{dionisio2004mutual} to limit the solution space during the evolutionary process, Bayesian information criterion (BIC) is used as a score function in with the genetic algorithm (GA) \cite{dai2020improved}.

The next example of a search space limiting algorithm is BiDAG which implements Markov chain Monte Carlo (MCMC) with a PC algorithm to reduce search space. The idea of this algorithm is to construct a Markov chain $M$, that has a stationary distribution equal to the posterior distribution $P(G|D)$ \cite{suter2021bayesian}.

\subsubsection{Local structures algorithms}

Another method for learning Bayesian networks is local structure algorithms. These algorithms learn local structures to construct the DAG. For example, Fei Liu and colleagues \cite{liu2016inference} propose an algorithm which can be described as:

\begin{enumerate}
    \item Generate an undirected graph, where features (nodes) are connected by edges if their MI is above a certain threshold;
    \item Split the generated graph into multiple networks using Markov blanket (MB) or kNN;
    \item The local Bayesian networks are learnt on these local graphs;
    \item Local networks are combined and new edges between networks are checked with conditional mutual information (CMI).
\end{enumerate}

Repeat clauses 2-4 until convergence \cite{liu2016inference}.

Another algorithm called deep-BN utilizes cliques of features. The algorithm is as follows: 

\begin{enumerate}
    \item A graph-based features clustering step puts the highly dependent features into similar groups. The particularity of our method consists of the
provided overlapped clusters, hence preserving as much information as possible.
    \item  Each found cluster of variables, in layer $l$, is represented by a corresponding latent variable in layer $l$+ 1. During this step, the cardinality of the latent variable is learnt based on the flow of information between the members of each cluster. Instead of the EM algorithm, the authors use a simplified implementation of the Equilibrium Criterion (EC) for learning the parameters of the latent variables.
\end{enumerate}

The stopping criterion of this iterative process induces the number of the hidden layer that will be included in the Deep-BN. The authors also provide a
stopping criterion for limiting the number of hidden layers, hence the information loss due to the insertion of the latent variables \cite{njah2019deep}.

\subsection{Existing packages for learning Bayesian networks with a large number of nodes}

Aragam and his co-authors suggest a $sparsebn$ R package \cite{Aragam2019}. The package has a large variety of tools from learning Bayesian networks to visualizing them. 

To learn Bayesian Networks $sparsebn$ package implements a score-based approach that uses a regularized maximum likelihood estimation. The main disadvantage of this package is the inability to work with mixed data when there are continuous and discrete variables in the data set. It also has a poor performance in terms of time on discrete data sets. \cite{aragamlearning}.

Another R library for building large Bayesian Networks is BiDAG. BiDAG is capable of working with discrete, continuous and mixed data. It also works with data sets with hundreds of features and has tools for dynamic Bayesian networks \cite{suter2021bayesian}.

Aforecited methods and algorithms are summarized in a table \ref{tab:table1}, their ability to accelerate different types of data, open-source implementation and size of data sets that were tested in the related article.
\begin{table}[H]
\centering
\caption{Related algorithms summary.}
\label{tab:table1}
\begin{tabular}{lllllll}
\hline
\multicolumn{1}{c}{\multirow{2}{*}{Article}} & \multicolumn{1}{c}{\multirow{2}{*}{\begin{tabular}[c]{@{}c@{}}Algorithms \\ and Methods\end{tabular}}} & \multicolumn{1}{c}{\multirow{2}{*}{\begin{tabular}[c]{@{}c@{}}Open-source\\ libriary\end{tabular}}} & \multicolumn{3}{c}{Data types acceleration}                                               & \multicolumn{1}{c}{\multirow{2}{*}{\begin{tabular}[c]{@{}c@{}}Size of tested\\ data sets\end{tabular}}}  \\ \cline{4-6}
\multicolumn{1}{c}{}                         & \multicolumn{1}{c}{}                                                                                   & \multicolumn{1}{c}{}                                                                                & \multicolumn{1}{c}{Discrete} & \multicolumn{1}{c}{Continuous} & \multicolumn{1}{c}{Mixed} & \multicolumn{1}{c}{}                                                                                     \\ \hline
\cite{fan2015improved}                       & MIIGA                                                                                                  & \xmark                                                                                              & \cmark                       & \xmark                         & \xmark                    & Up to 100 (Alarm)                                                                                        \\
\cite{liu2016inference}                      & CMI, MI, kNN                                                                                           & \xmark                                                                                              & \cmark                       & \xmark                         & \xmark                    & Up to 100                                                                                                \\
\cite{suter2021bayesian}                     & \begin{tabular}[c]{@{}l@{}}MCMC\\ with PC\end{tabular}                                                 & \begin{tabular}[c]{@{}l@{}}$BiDAG$ R\\ package\end{tabular}                                         & \cmark                       & \cmark                         & \cmark                    & \begin{tabular}[c]{@{}l@{}}Hundreds of\\ nodes (gsim100, \\ gsim)\end{tabular}                           \\
\cite{njah2019deep}                          & deep-BN                                                                                                & \xmark                                                                                              & \cmark                       & \cmark                         & \cmark                    & Up to 312 nodes                                                                                          \\
\cite{aragamlearning}                        & \begin{tabular}[c]{@{}l@{}}Regularized\\ maximum\\ likelihood\end{tabular}                             & \begin{tabular}[c]{@{}l@{}}$sparsebn$ R\\ package\end{tabular}                                      & \cmark                       & \cmark                         & \xmark                    & \begin{tabular}[c]{@{}l@{}}Thousands of nodes,\\ data sets with\\ above 8000 nodes\\ tested\end{tabular} \\ \hline
\end{tabular}
\end{table}

According to the table \ref{tab:table1} only $BiDAG$ R package meets all the mentioned parameters. The rest of the algorithms either do not work with continuous or mixed data or lack open-source implementation.

\section{Background}
\label{bg}

A Bayesian network is a directed acyclic graph (DAG) in which the nodes ($V$) represent the features being modelled, and the directed edges ($E$) indicate the presence of dependencies between the features. Thus, a multivariate distribution $P(X_1,\dots,X_p)$ can be represented as a product of conditional distributions, where the conditional distributions are determined by the structure of the Bayesian network:
\begin{equation}
    P(X_1,\dots,X_p )=\prod_{j=1}^p P(X_j |\Pi_{X_j}).
\end{equation}

The task of learning the structure of a Bayesian network can be formulated as an optimization problem:
\begin{equation}
V_{opt}, E_{opt} = \operatorname*{argmax}_{G'\in G_{possible}} F(G'),   
\end{equation}
where $V_{opt}, E_{opt}$ - nodes and edges of the found optimal BN; $G_{possible}$ - the search space of all possible structures of BN; $F$ - score function which indicates how well the structures fits data. Here we can use different score functions, for example, K2 \cite{behjati2020improved}, BIC \cite{scanagatta2019survey} and etc. However, with a large number of nodes, the space of possible structures $G_{possible}$ becomes too large, therefore, it needs to be limited. If we measure the strength of links between features, then the space of possible structures can be reduced by excluding edges between features, the strength of the connection between which is below some given threshold:
\begin{equation}
G_{new} = {G : G \in G_{possible} \land Strength(Edges(G)) > Threshold}
\end{equation}
The advantage of this approach is that by varying the threshold value, we can control the size of the search space and thus obtain different time estimates depending on the desired result.

Thus, our goal is to develop an algorithm that, using a thresholding approach, finds optimal structures for Bayesian networks with a large number of nodes (more than 50) without loss of quality and in a reasonable time.

\section{Algorithm}
\label{alg}

\begin{figure}[H]
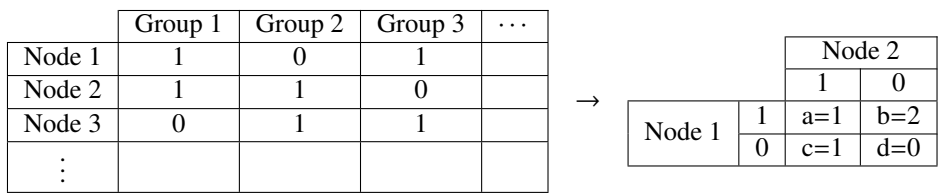

    \centering
    \begin{tabular}{ | *{5}{ c | } }
        \cline{2-5}
        \multicolumn{1}{c|}{} & Group 1 & Group 2 & Group 3 & $\cdots$\\\hline
        Node 1 & 1 & 0 & 1 &  \\\hline
        Node 2 & 1 & 1 & 0 &  \\\hline
        Node 3 & 0 & 1 & 1 &  \\\hline
        $\vdots$ & & & & \\\hline
    \end{tabular}\hspace{1em}$\rightarrow$\hspace{1em}\begin{tabular}{ | *{4}{ c | } }
        \cline{3-4}
        \multicolumn{2}{c|}{} & \multicolumn{2}{c|}{Node 2}\\\cline{3-4}
        \multicolumn{2}{c|}{} & 1 & 0 \\\hline
        \multirow{2}{*}{Node 1} & 1 & a=1 & b=2 \\\cline{2-4}
        & 0 & c=1 & d=0 \\\hline
    \end{tabular}\\\vspace{1em}
    \caption{Brave coefficient calculation method.}
    \label{fig:nodetransform}
\end{figure}

The $BigBraveBN$ has been built using BAMT library\footnote{BAMT, Repository experiments and data, \href{https://github.com/ITMO-NSS-team/BAMT.git}{https://github.com/ITMO-NSS-team/BAMT.git}, 2021}, that was developed by our laboratory team. As an algorithm for finding the optimal structure of BN, we chose the greedy Hill-Climbing algorithm with the K2 score function, which we have already implemented in the BAMT library. The idea of this algorithm is quite simple, we start searching for a structure from an empty graph and add, delete or reverse one edge at each iteration, and if the value of the score function improves, we fix this action with an edge. To limit the search space, we propose our algorithm based on the Brave coefficient \cite{degteva2009system}. This coefficient measures the mutual occurrence of variables when clustering is applied to the data set. The algorithm of Brave coefficient calculation is shown in Figure \ref{fig:nodetransform}. In the first step, we initialize a matrix that represents $n$ nearest neighbours for every variable (groups). In the second step for every pair of variables Brave coefficient is calculated using formula \ref{equation:Br}. In this formula, $a$ is the number of groups in which both features fell, $b$ and $c$ are the numbers of groups in which one feature fell, but the second did not fall, $d$ is the number of groups in which none of the features fell, $n$ - dataset size.

\begin{equation}
    \label{equation:Br}
	Br = \frac{a \times n + (a + c) \times (a + b)}{\sqrt{(a + c) \times (b + d)} + \sqrt{(a + b) \times (c + d)}}
\end{equation}

\begin{figure}[H]
\centering
\begin{tikzpicture}
	\node [label={Data set}] (dataset) at (0,0) {
		\begin{tabular}{ | *{5}{ c | } }
		\cline{2-5}
		\multicolumn{1}{c|}{} & A & B & C & $\cdots$ \\\hline
		1 & 1 & 2 & 3 & \\\hline
		2 & 0 & 4 & 5 & \\\hline
		3 & 1 & 3 & 4 & \\\hline
		$\vdots$ & & & & \\\hline
		\end{tabular}
	};
	\node (proxmeas) at (4,0) {
		\parbox{2cm}{
			\centering
			Proximity measure of variables
		}
	};
	\node [label={Proximity matrix}] (proxmatrix) at (8,0) {
		\begin{tabular}{ *{6}{ c | } }
			\cline{2-6}
			A &&&&& \\\cline{2-6}
			B &&&&& \\\cline{2-6}
			$\vdots$ &&&&& \\\cline{2-6}
			Y &&&&& \\\cline{2-6}
			Z &&&&& \\\cline{2-6}
			\multicolumn{1}{c}{} & \multicolumn{1}{c}{A} & \multicolumn{1}{c}{B} & \multicolumn{1}{c}{$\cdots$} & \multicolumn{1}{c}{Y} & \multicolumn{1}{c}{Z}
		\end{tabular}
	};
	\node (nearest) at (8,-4) {
		\parbox{2cm}{
			\centering
			Get N nearest neighbors by proximity
		}
	};
	\node (brave) at (4,-4) {
		\parbox{2cm}{
			\centering
			Calculate Brave coefficient
		}
	};
	\node [label={Brave matrix}] (brmatrix) at (0,-4) {
		\begin{tabular}{ *{6}{ c | } }
			\cline{2-6}
			A &&&&& \\\cline{2-6}
			B &&&&& \\\cline{2-6}
			$\vdots$ &&&&& \\\cline{2-6}
			Y &&&&& \\\cline{2-6}
			Z &&&&& \\\cline{2-6}
			\multicolumn{1}{c}{} & \multicolumn{1}{c}{A} & \multicolumn{1}{c}{B} & \multicolumn{1}{c}{$\cdots$} & \multicolumn{1}{c}{Y} & \multicolumn{1}{c}{Z}
		\end{tabular}
	};
	\node (whlist) at (0,-7) {
		\parbox{2cm}{
			\centering
			White list by threshold
		}
	};
	\node (end) at (4,-7) {
		\parbox{3cm}{
			\centering
			Structure learning via HillClimbing in BAMT
		}
	};
	\draw [->] (dataset)	edge (proxmeas)
		(proxmeas) edge (proxmatrix)
		(proxmatrix) edge (nearest)
		(nearest) edge (brave)
		(brave) edge (brmatrix)
		(brmatrix) edge (whlist)
		(whlist) edge (end)
	;
\end{tikzpicture}
\caption{BigBraveBN algorithm workflow.}
\label{fig:dataset}
\end{figure}
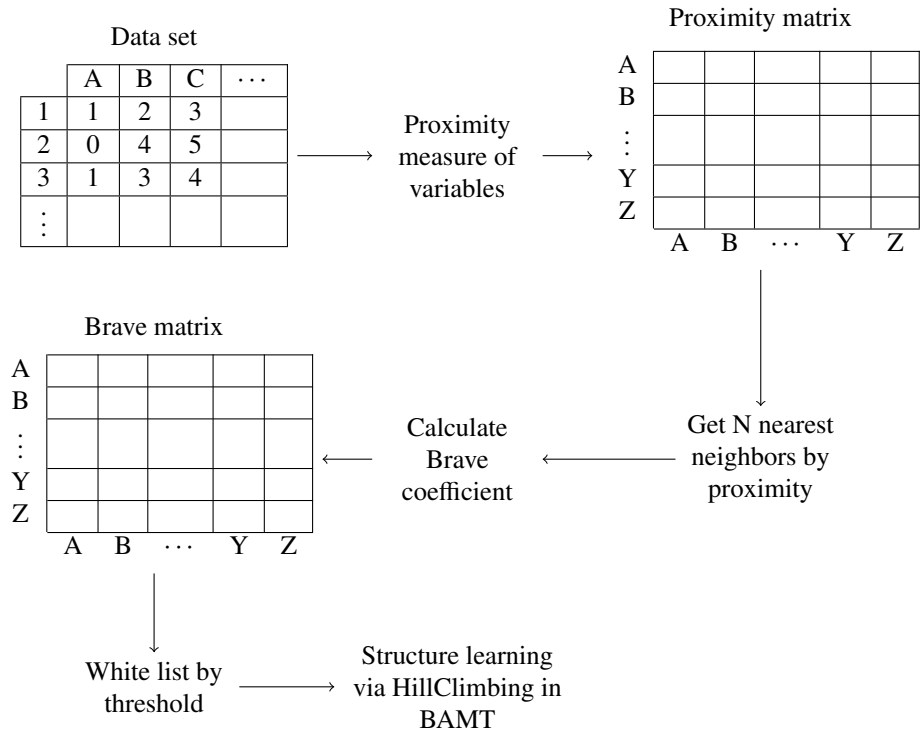

Figure \ref{fig:dataset} represents a workflow of the $BigBraveBN$ algorithm. At this point two main hyperparameters are available: threshold and number of nearest neighbors for each variable.  

\begin{itemize}
    \item \textbf{Step 1} Initialize data set;
    \item \textbf{Step 2} Choose a proper proximity measure;
    \item \textbf{Step 3} Apply proximity measure to the data set;
    \item \textbf{Step 4} Get $N$ nearest neighbours by proximity for every variable;
    \item \textbf{Step 5} Calculate Brave coefficient of mutual occurrence on nearest neighbors;
    \item \textbf{Step 6} Get Brave coefficient matrix;
    \item \textbf{Step 7} Generate white list of possible edges by setting a threshold;
    \item \textbf{Step 8} Perform structure learning via BAMT tools.
\end{itemize}

To form N-nearest neighbours different proximity measures can be used. But some of them have certain limitations. Pearson correlation coefficient \cite{benesty2009pearson} and MI \cite{dionisio2004mutual} metrics were reviewed for being used in the algorithm. MI metric was chosen as a default metric due to its versatility and efficiency.

Here, when we run the Hill-Climbing algorithm, we limit the search space by giving the algorithm a so-called white list. The white list is a predefined manually or by a certain algorithm list of edges, that limits the solution space by restricting any other connection except itemized in the white list. Thus, the Bayesian Network can only consist of edges that are included in a specific white list.
\section{Experiments}
\label{exp}

\subsection{BigBraveBN hyperparameters tests}

Since $BigBraveBN$ provides some hyperparameters, it is essential to test which values lead to a decent execution time and meet certain requirements. To measure the time and SHD dynamics over different hyperparameters, pigs data set \cite{bnlearn} was chosen. Pigs data set has 441 features (nodes) and its reference structure has 592 edges. 

All the experiments in the experimental part of the paper were carried out on an AMD Ryzen™ 7 5800X CPU.

\begin{figure}[H]\vspace*{4pt}
\label{fig:hptests}
\centerline{\includegraphics[scale=.3]{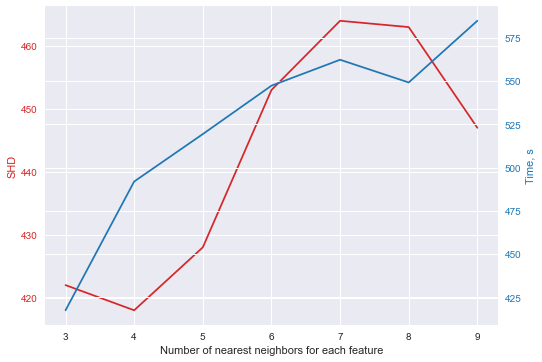}\hspace*{5mm}\includegraphics[scale=.3]{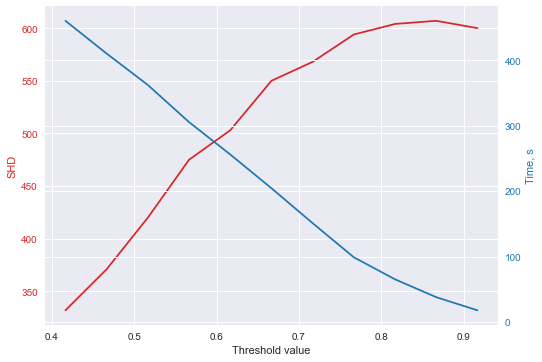}}
\caption{(a) Number of nearest neighbours test of $BigBraveBN$, pigs data set; (b) Threshold value test of $BigBraveBN$, pigs dataset.}
\end{figure}

Figure \ref{fig:hptests} represents Structural Hamming Distance (SHD) and execution time dynamics over two hyperparameters: the number of nearest neighbours and the threshold value of the Brave coefficient. Structural Hamming Distance is several deleted, reversed or added edges to transform one graph into another \cite{peters2015structural}. The less SHD the more precise the structure is when compared to the true structure. Thus, it is necessary to choose hyperparameters that meet certain requirements such as acceptable SHD and execution time. 

According to hyperparameters tests on different data sets, the number of nearest neighbours is chosen equal to 5, the threshold is equal to $(0.3 \cdot Br_{max})$, where $Br_{max}$ is the highest Brave coefficient of the Brave coefficient matrix.

\subsection{General comparison in terms of Structural Hamming Distance}
\label{generalSHD}

In this section $BigBraveBN$ is compared to $sparsebn$ \cite{Aragam2019} and $BiDAG$ \cite{suter2021bayesian} R packages. Comparison is based on two parameters: time of structure learning and precision represented by SHD. 

There are very important features and disadvantages regarding the R packages mentioned above that should be discussed before presenting the results. First of all $sparsebn$ performs quite outstanding on data sets where all the variables are continuous, however performance on data sets with discrete variables is poor in terms of time, thus only $BigBraveBN$ and $BiDAG$ will be compared on discrete data sets in terms of time.

\begin{figure}[H]
    \centering
    \includegraphics[scale = 0.2]{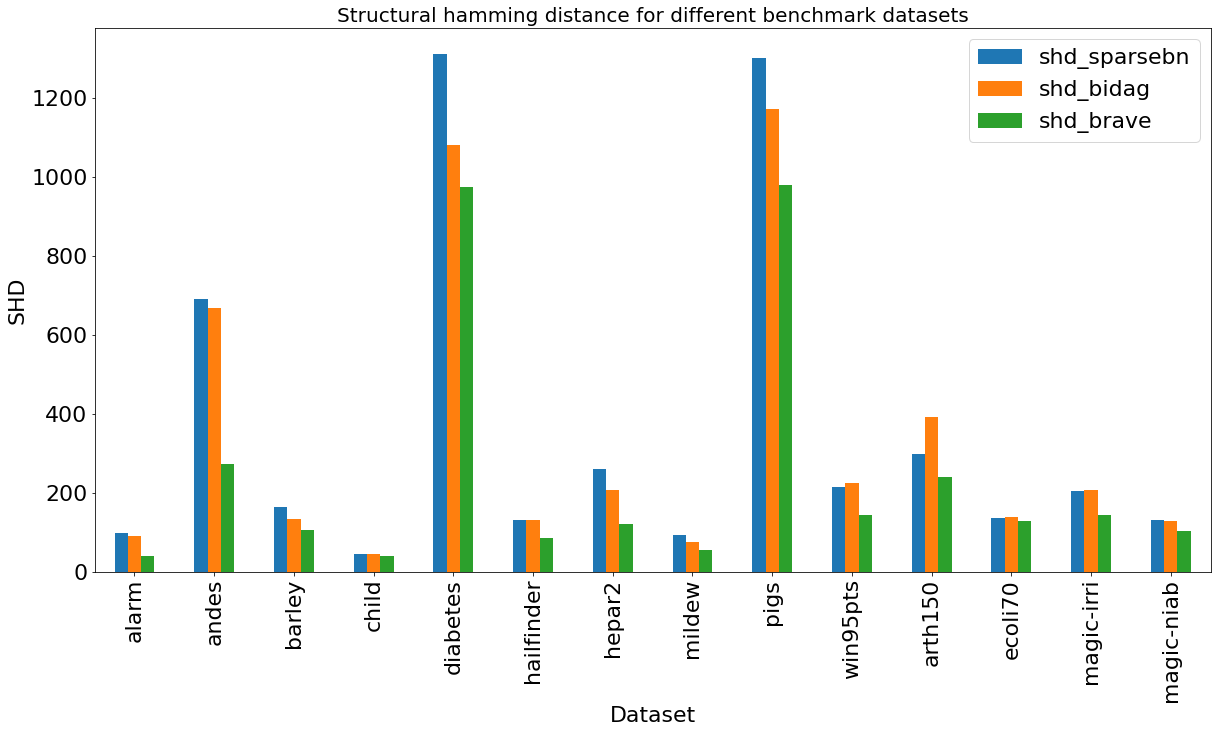}
    \caption{Structural Hamming Distance on different benchmark data sets from bnlearn.}
    \label{fig:SHD_all}
\end{figure}

All the data sets mentioned on the plots below have a reference structure, presented on $bnlearn$ \footnote{bnlearn - an R package for Bayesian network learning and inference, Bayesian Network Repository \href{https://www.bnlearn.com/bnrepository/}{https://www.bnlearn.com/bnrepository/}} website. Thus, all the obtained by $BigBraveBN$, $sparsebn$ and $BiDAG$ algorithms were compared to the $bnlearn$ reference structures in terms of SHD on Figure \ref{fig:SHD_all}. Presented data sets have from 20 up to 441 nodes.

\subsection{Comparison of BigBraveBN with randomly chosen edges in terms of Structural Hamming Distance}
\label{random}

The next essential experiment comes down to a comparison of $BigBraveBN$ SHD and SHD obtained by a randomly generated set of edges of the same size. This experiment is designed to test the consistency of the proposed algorithm and show that the algorithm reduces the search space by identifying related features.

\begin{figure}[H]
    \centering
    \includegraphics[scale = 0.18]{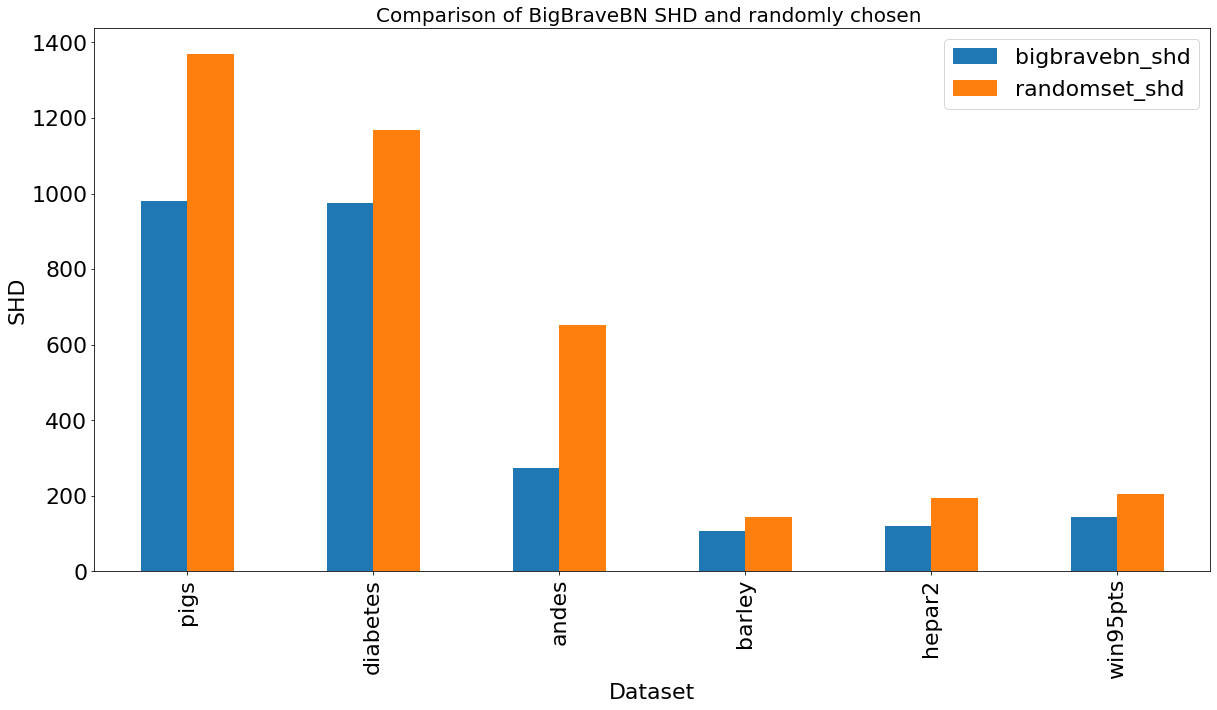}
    \caption{Structural Hamming Distance of BigBraveBN compared to random choice of edges on different benchmark data sets.}
    \label{fig:SHD_random}
\end{figure}

Figure \ref{fig:SHD_random} shows, that the same number of edges, but taken randomly has higher SHD, compared to $BigBraveBN$. This suggests that a random reduction in the search space due to a random selection of edges will not lead to a qualitative result, which means that the proposed algorithm is indeed capable of reducing the search space by identifying strongly related features.

\subsection{Discrete data sets, time comparison}
\label{disctime}

\begin{figure}[H]
    \centering
    \includegraphics[scale = 0.18]{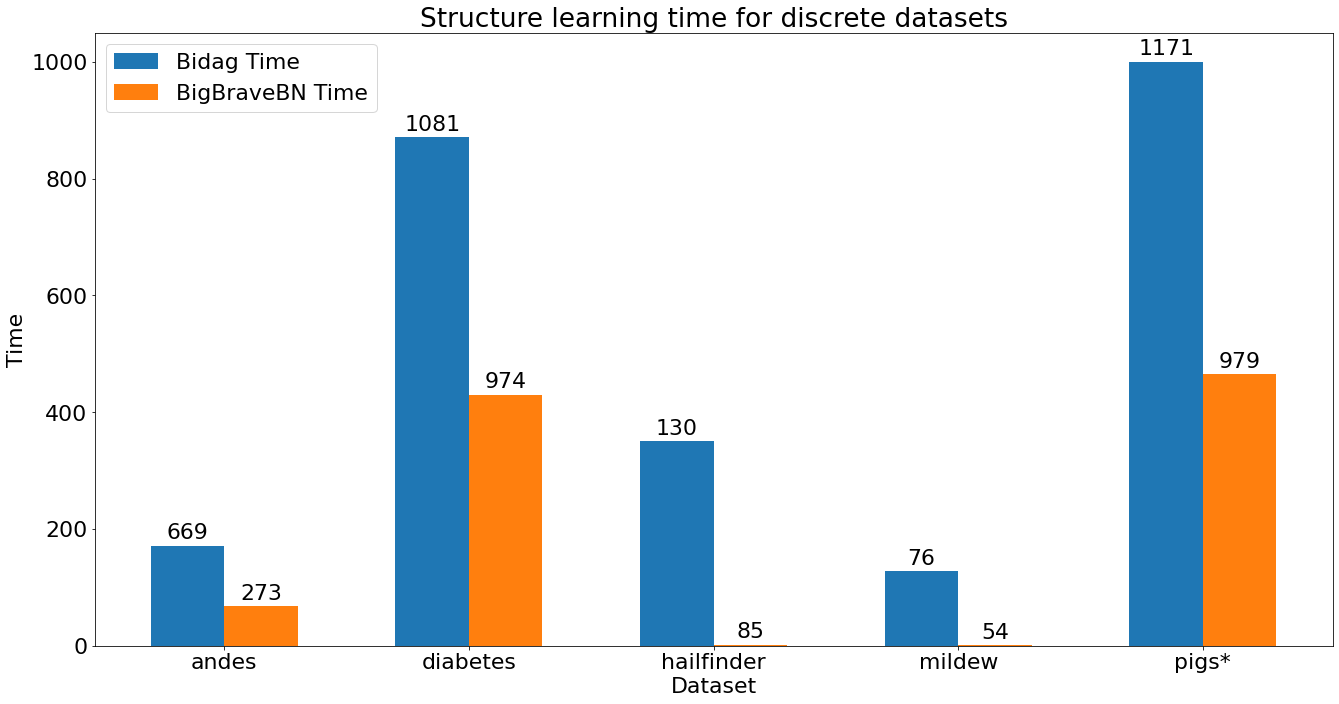}
    \caption{Structural learning time in seconds for bigger discrete benchmark data sets time, taken to learn structure of pigs data (441 nodes) set by $BiDAG$ is over eleven thousands seconds.}
    \label{fig:time_comp_disc}
\end{figure}

In the case of discrete data sets in terms of time, $BigBraveBN$ is compared to $BiDAG$, because of the two mentioned above R packages, it performs significantly faster on discrete data sets. However, $BigBraveBN$ performs even faster as shown on Figure \ref{fig:time_comp_disc}. Combined with a lower SHD score (numbers above the columns in the Figure \ref{fig:time_comp_disc}) these results show, that $BigBraveBN$ algorithm has better time performance and precision. 

\subsection{Continuous data sets, time comparison}
\label{conttime}

\begin{figure}[H]
    \centering
    \includegraphics[scale = 0.18]{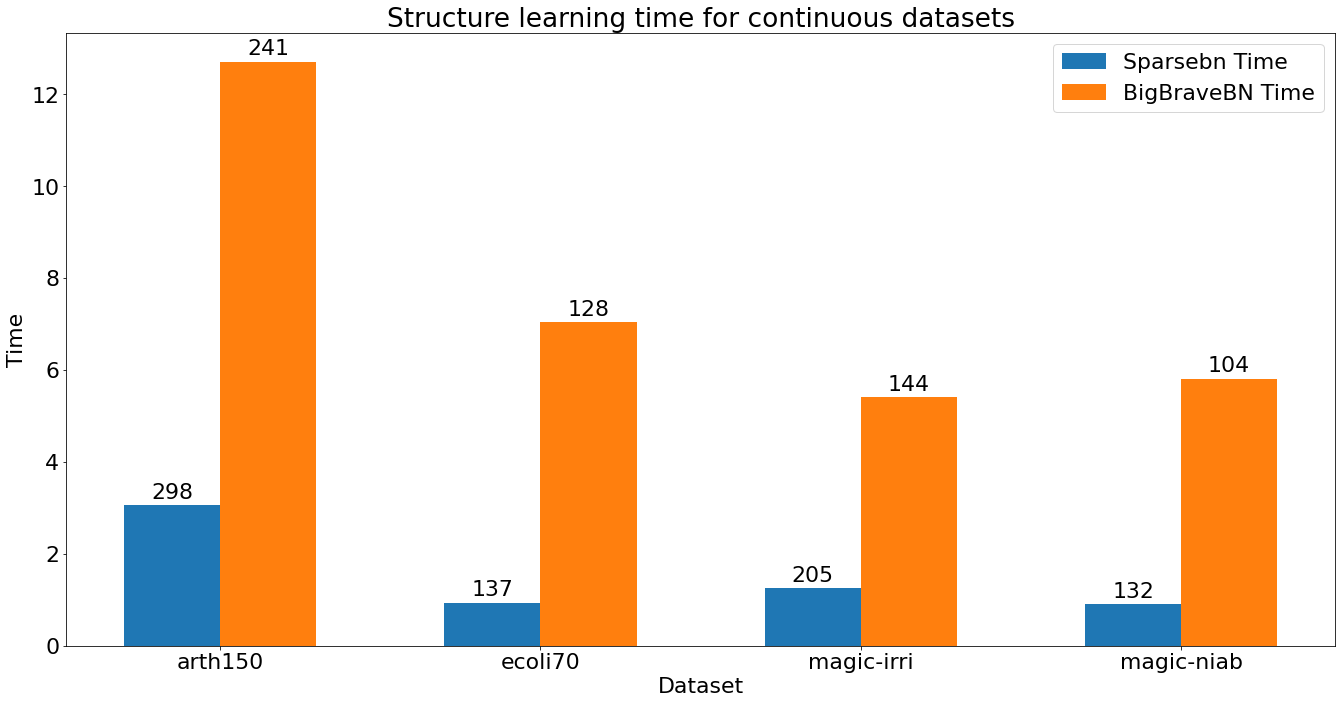}
    \caption{Structural learning time in seconds for bigger continuous benchmark data sets.}
    \label{fig:time_comp_cont}
\end{figure}

In the case of continuous data, $BigBraveBN$ is compared to $sparsebn$, because it is significantly faster than $BiDAG$ on continuous data. In case of all the data sets $sparsebn$ performs structural learning faster (Figure \ref{fig:time_comp_cont}), but in terms of SHD $BigBraveBN$ still shows better results (numbers above the columns in the Figure \ref{fig:time_comp_cont}), however, the increase in time is negligible.

\subsection{$BigBraveBN$ time complexity}
\label{Complexity}

To define time complexity depending on several variables the following experiment was implemented. From both the largest discrete and largest continuous data sets (in terms of the number of variables) a certain number (20, 40, ... N, where N is the full number of variables of the data set) of variables were randomly taken. And on these groups of variables, the structure learning was performed by $BigBraveBN$, $BiDAG$ and $sparsebn$. 

As expected, the time complexity of $BigBraveBN$ is better than the complexity of $BiDAG$ on discrete data, however, $BigBraveBN$ loses in terms of complexity to $sparsebn$ on continuous data, but not dramatically. 

\begin{figure}[H]\vspace*{4pt}
\centerline{\includegraphics[scale=.3]{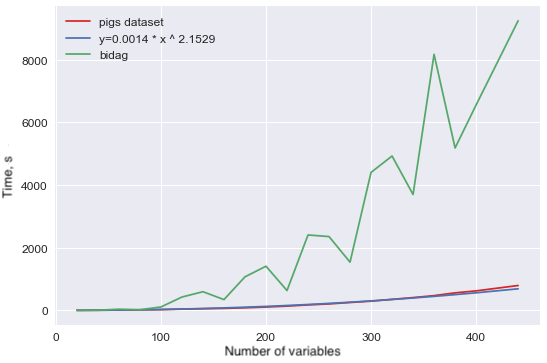}\hspace*{5mm}\includegraphics[scale=.3]{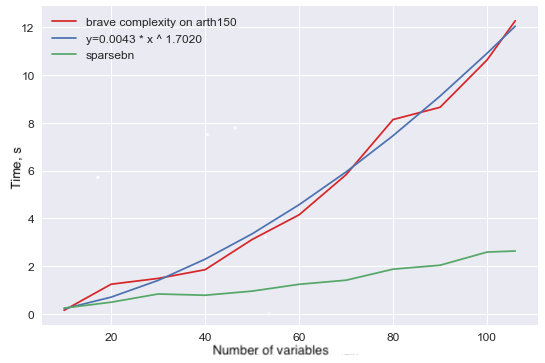}}
\caption{(a) Time complexity of $BigBraveBN$ on discrete data set (red line) compared to $BiDAG$ complexity (green line); (b) Time complexity of $BigBraveBN$ on continuous data set (red line) compared to $sparsebn$ complexity (green line).}
\end{figure}

\subsection{Experiments on real data}

To test the ability of the proposed algorithm to work with data from real-world objects, the following experiment was carried out. This algorithm can be used to build Bayesian networks based on the real anonymized medical data from residents of St. Petersburg, or any other social, economic, and other data.

For the experiment, a medical dataset was chosen. The dataset includes medical records of 880 patients with 56 features both continuous and discrete treated for type 2 diabetes mellitus at the N.N. Almazov and the First St. Petersburg State Medical University. Pavlov, St. Petersburg, Russia, in 2008-2018 \cite{pavlovskii2021hybrid}.
\begin{figure}[H]
  \centering
  \includegraphics[scale=.15]{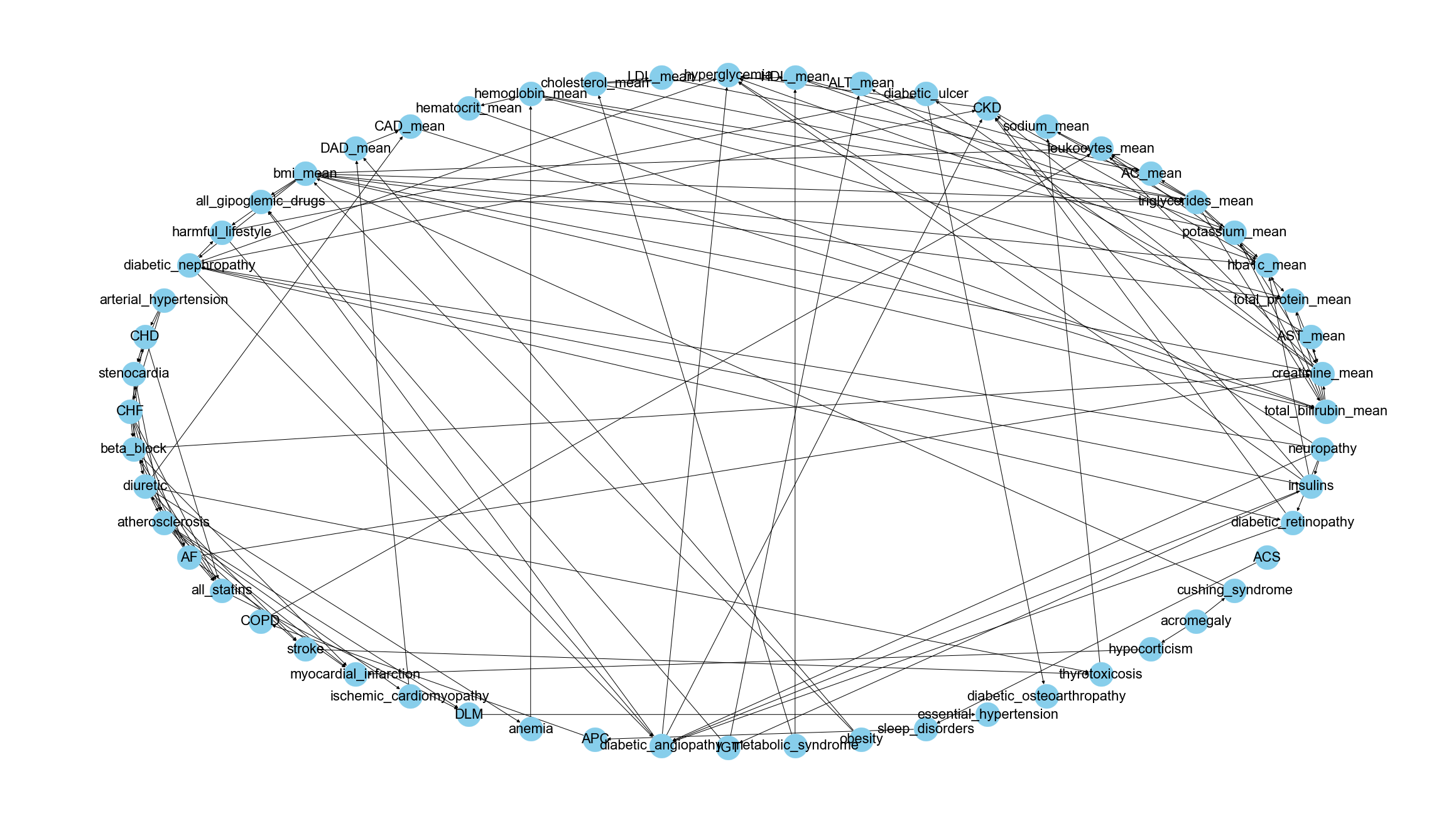}
  \caption{Bayesian Network structure represented as DAG obtained by $BigBraveBN$ algorithm}
  \label{fig:medicalbn}
\end{figure}

\begin{figure}[htp]\vspace*{4pt}
    \centerline{\includegraphics[scale=.15]{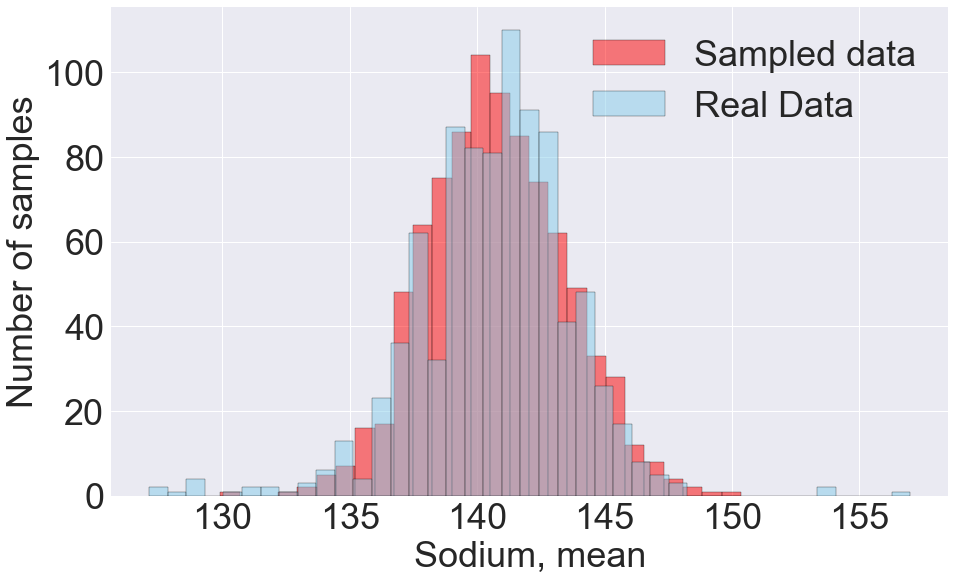}\hspace*{5mm}\includegraphics[scale=.15]{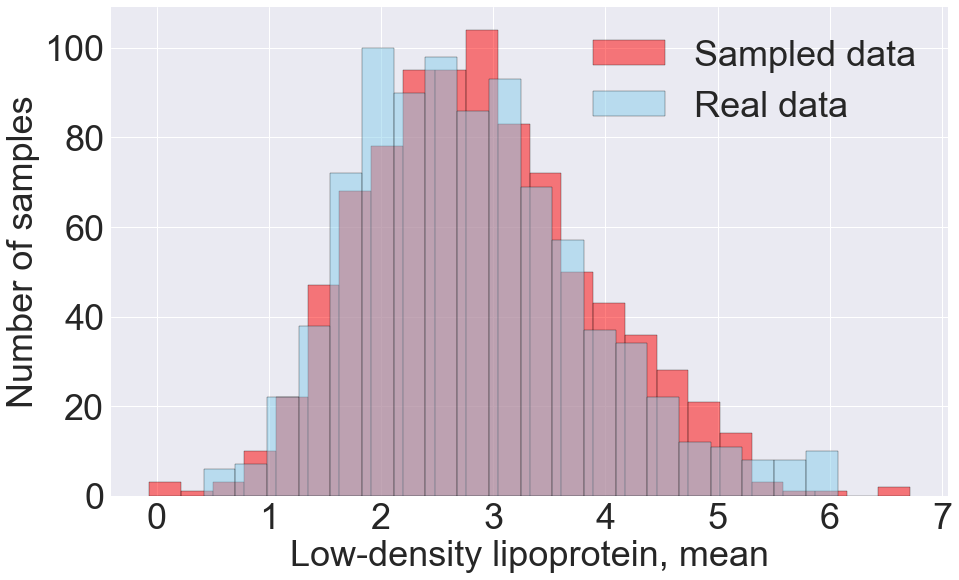}\hspace*{10mm}\includegraphics[scale=.15]{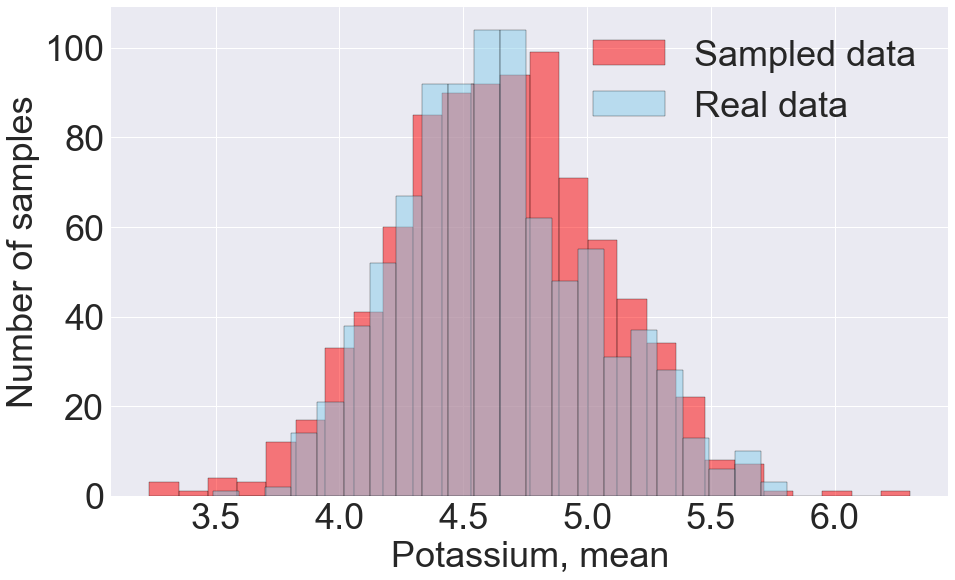}}
    \caption{Sampling distribution of (a) mean Sodium; (b) mean Low-density lipoproteins; (c) mean Potassium}
    \label{fig:sampling}
\end{figure}

Barplots in the figure \ref{fig:sampling} show the distribution of continuous real and sampled data using the Bayesian network shown in figure \ref{fig:medicalbn}, built by the $BigBraveBN$ algorithm. According to the barplots, the sampling is quite accurate which indicates a good quality of simulation, although the training time of the structure took only seven seconds.

\section{Conclusion and Discussion}
\label{conclusion}

The article considered the problems of structural learning of Bayesian networks with a large number of nodes. As a result of resolving the shortcomings of existing learning algorithms, our own algorithm $BigBraveBN$ was proposed. The $BigBraveBN$ algorithm implements an outstanding tool for structure learning of large Bayesian networks on discrete, continuous and mixed types of data. The algorithm is based on a Brave coefficient that utilizes the N-nearest neighbours approach based on the MI metric. In the experimental part, we have shown that our algorithm has a decent efficiency on both synthetic and real data. The algorithm was compared with the existing analogues of $sparsebn$ and $BiDAG$. As a result of the comparison, the proposed algorithm not only worked faster but also showed higher efficiency in terms of the accuracy of restoring BN structures.

Shortly, we plan to integrate $BigBraveBN$ into the BAMT library. $BigBraveBN$ can be also used for structure learning not only via Hill-Climbing but also with help of genetic algorithms. Thus, our future work is also concentrated on a fusion of $BigBraveBN$ with genetic algorithms.

Software implementation of the algorithm and experimental results can be found in the repository \cite{BigBraveBN}.
\bibliography{literature}
\bibliographystyle{elsarticle-harv}

\end{document}